\documentclass[10pt,journal,final]{IEEEtran}
\usepackage[T1]{fontenc}

\usepackage{graphicx}
\graphicspath{{./Figures/}}

\usepackage{amsmath}
\usepackage{amssymb}
\usepackage{bm}
\usepackage{algorithmic}  
\usepackage{algorithm}  
\usepackage{booktabs}
\usepackage{cite}
\usepackage[colorlinks,linkcolor=black, citecolor=black, urlcolor=black]{hyperref}

\ifCLASSOPTIONcompsoc
    \usepackage[caption=false, font=normalsize, labelfont=sf, textfont=sf]{subfig}
\else
\usepackage[caption=false, font=footnotesize]{subfig}
\fi

\begin{document}
\title{Privacy-preserving Spatiotemporal Scenario Generation of Renewable Energies: A Federated Deep Generative Learning Approach}

\author{Yang~Li,~\IEEEmembership{Senior Member,~IEEE,} Jiazheng~Li and Yi~Wang,~\IEEEmembership{Member,~IEEE}%
\thanks{This work was supported in part by the National Natural Science Foundation of China-State Grid Corporation of China Smart Grid Joint Fund Key Program (U2066208), and the Natural Science Foundation of Jilin Province, China (2020122349JC). Paper no. TII-21-0365. (Corresponding author: Yang~Li and Yi~Wang.)}%
\thanks{Y.~Li, J.~Li are with the School of Electrical Engineering, Northeast Electric Power University, Jilin 132012, China (e-mail: liyang@neepu.edu.cn; tytrytytgr@gmail.com).}%
\thanks{Y.~Wang is with the Power Systems Laboratory, Department of Information Technology and Electrical Engineering, ETH Zürich, 8092 Zürich, Switzerland (e-mail: yiwang@eeh.ee.ethz.ch)}
}

\maketitle

\begin{abstract}
    Scenario generation is a fundamental and crucial tool for decision-making in power systems with high-penetration renewables. Based on big historical data, a novel federated deep generative learning framework, called Fed-LSGAN, is proposed by integrating federated learning and least square generative adversarial networks (LSGANs) for renewable scenario generation. Specifically, federated learning learns a shared global model in a central server from renewable sites at network edges, which enables the Fed-LSGAN to generate scenarios in a privacy-preserving manner without sacrificing the generation quality by transferring model parameters, rather than all data. Meanwhile, the LSGANs-based deep generative model generates scenarios that conform to the distribution of historical data through fully capturing the spatial-temporal characteristics of renewable powers, which  leverages the  least squares loss function to improve the training stability and generation quality. The simulation results demonstrate that the proposal manages to generate high-quality renewable scenarios and outperforms the state-of-the-art centralized methods. Besides, an experiment with different federated learning settings is designed and conducted to verify the robustness of our method.
\end{abstract}
    

\section{Introduction}
\IEEEPARstart{T}{he} development and utilization of renewable energies is an inevitable choice to solve environmental pollution and ensure sustainable energy supply. However, due to its inherent uncertainty and intermittency, the integration of renewable energies has brought huge challenges to the planning and operation of today's power systems \cite{dolatabadi2017optimal}. At present, how to handle the uncertainty of renewable generations is currently the difficulty of studying the economic optimization of power systems. Scenario analysis is a technique to characterize the renewable energy generation by generating a set of typical time series scenarios that conform to the statistical characteristics of the renewable energies power outputs, thereby transforming a uncertainty optimization problem with renewables into a deterministic one. By using the generated renewable power scenarios, power system operating personnels are able to make reasonable decisions taking into account the uncertainty of renewable generation outputs\cite{lahon2018optimal}. 

\subsection{Literature review}
There have been a number of studies carried out on different aspects of renewable scenario generation. Unfortunately, most of previous works focus on model-based scenario generation methods, which must know the probabilistic distributions of renewable outputs. For example, the work in \cite{talari2017optimal} generates wind power scenarios based on Monte Carlo method and using constructed Rayleigh distribution. In~\cite{papaefthymiou2008using}, a framework using Gaussian copula is presented for investigating the random dependence scenarios of wind power in a multivariate context.  Reference~\cite{ma2013scenario}, a empirical cumulative distribution function is applied to model wind power uncertainty and then generate plenty of scenarios through the inverse transform sampling of the multivariate normal distribution.  In~\cite{lee2016load}, the generalized dynamic factor model (GDFM) is adopted to generate a synthesized scenario of load and wind powers, where GDFM retains the correlation structure between load and wind powers. However, the above-mentioned model-based methods suffer from some open problems, such as the requirement for a priori assumption, trivial modeling processes and low quality of generated scenarios.

With the development of artificial intelligence, emerging deep generative models (DGMs) have recently received a great deal of attention for applying to renewable power scenario generation. As a powerful DGM, generative adversarial networks (GANs) can learn the latent distributions of real data and generate real-like data in an unsupervised manner without explicit modeling. GANs use the two player zero-sum game to simultaneously train two deep neural networks: generator and discriminator. The generator network converts the noise distribution into artificial data by learning the distribution mapping relationship of known historical data, and the discriminator is trained to differentiate between generated and historical data~\cite{goodfellow2014generative}. Reference~\cite{chen2018model} is the first work that utilizes GANs to generate renewable power scenarios. Sequence GANs~\cite{liang2019sequence} and improved Wasserstein GANs~\cite{zhang2020typical} are also employed for performing such tasks. Moreover, GANs-based scenario generation methods have been used in some optimal scheduling studies, and the results show that this kind of methods are conducive to the formulation of more secure and economical power system operation strategies because GANs is able to generate high-quality scenarios with spatio-temporal correlation~\cite{kong2020robust,wei2019short}.

However, all the above existing methods need to transmit all data of each wind/solar power plant to a data center with high-performance computing clusters for generating renewable scenarios. Such kind of centralized data handling approaches will suffer from the following problems: 1) these approaches depend on a central workstation with powerful computing power and storage space; 2) frequent data exchanges between renewable energy plants and a central workstation will inevitably cause heavy communication overheads; 3) data stored in a centralized form is vulnerable to data leakage and cyber-attacks; 4) most importantly, some renewable energy data owners, such as independent system operators (ISOs), are not willing to share data with others in practical applications due to data privacy. In this context, data privacy can refer to either commercially sensitive data from grid-connected renewable energy plants with commercial competition or personal data from households with renewable energy technology\cite{goncalves2021privacy}. At present, many countries have issued special laws such as the General Data Protection Regulation (GDPR)~\cite{GDPR} enforced by the European Union and the China's Cyber Security Law and the General Principles of the Civil Law to regulate the management and use of data. Consequently, there is an urgent need to find a distributed scenario generation method that can solve the above problems. 

Some distributed structures considering data security and privacy protection have been studied in~\cite{jia2018distributed, goncalves2021privacy, ferdowsi2020brainstorming}. Reference~\cite{goncalves2021privacy} formulates a  privacy-preserving framework that combines data transformation techniques with the alternating direction method of multipliers for renewable energy forecasting. In \cite{jia2018distributed}, a distributed maximum a posteriori probability estimation method is developed to build a Gaussian mixture model of aggregated wind power forecast error for addressing the data privacy problem. Regarding distributed DGM, reference~\cite{ferdowsi2020brainstorming} proposes a brainstorming GAN (BGAN) for generating data when multiple agents are unwilling to share data due to privacy issues. However, the BGAN architecture is not suitable for renewable scenario generation, because each agent of BGAN generates samples not only similar to their own dataset, but also similar to their neighbors' dataset, which will  affect the decision-making of operators.

In this context, federated learning, as an extension of distributed machine learning, has become an increasingly hot topic in recent years~\cite{li2020federated}, as it provides a new way of model learning over a collection of highly distributed devices while preserving data privacy and reducing communication overhead through the decoupling of model training from direct access to original training dataset. Federated learning has been successfully applied in various fields such as electricity consumer characteristics identification~\cite{wang2021electricity}, solar irradiation forecasting~\cite{zhang2020probabilistic} and energy management of multiple smart homes with a photovoltaic system~\cite{lee2020federated}. This paper introduces federated learning into the field of scenarios generation for the first time, which enables each local generation model to collaboratively learn a shared global model while preserving data locality during training.

\subsection{Contribution of This Paper}
In this paper, a brand new federated deep generative learning based renewable energy scenario generation approach that integrates with federated learning and least squares generative adversarial networks (LSGANs), called Fed-LSGAN, is proposed. The main contributions of this paper are summarized as the following three folds:

1) This paper proposes Fed-LSGAN, a novel federated learning-based framework for  renewable scenario generation. To the authors' knowledge, this is the first work to leverage federated learning for scenario generation. Here, federated learning learns a shared global model in a central server from renewable power plants at network edges, which enables the Fed-LSGAN to generate scenarios in a privacy-preserving  manner without sacrificing the generation quality by transferring model parameters, rather than all data.

2) The Fed-LSGAN features on a new design of the LSGANs-based deep generative model in a distributed fashion for generating scenarios that conform to the distribution of historical data through fully capturing the temporal and spatial characteristics of renewable powers, which leverages the least squares loss function to improve the training stability and the quality of generated scenarios.

3) Extensive tests have been performed using a real-world dataset to examine the effectiveness of the Fed-LSGAN, which demonstrates that Fed-LSGAN is superior to state-of-the-art centralized methods. Besides, an experiment with different control parameters in federated settings is designed to examine the robustness of our method.

\subsection{Organization of This Paper}
This paper is organized as follows. Section~\ref{sec.related_work} introduces the basic concept of GANs and federated learning. Section~\ref{sec.fedgan} proposes a distributed scenario generation model based on federated learning mechanism. Section~\ref{sec.case_study} applies the proposed model to real-world data sets, and verifies the advantages of the model. Section~\ref{sec.conclusion} concludes this paper.

\section{Related Works}\label{sec.related_work}
To elicit the method used in this article, the principle and formula of GANs are firstly introduced, and then the  federated learning mechanism is described in detail.

\subsection{Generative Adversarial Networks}
\begin{figure}
	\centering	
	\includegraphics[width=3.4in]{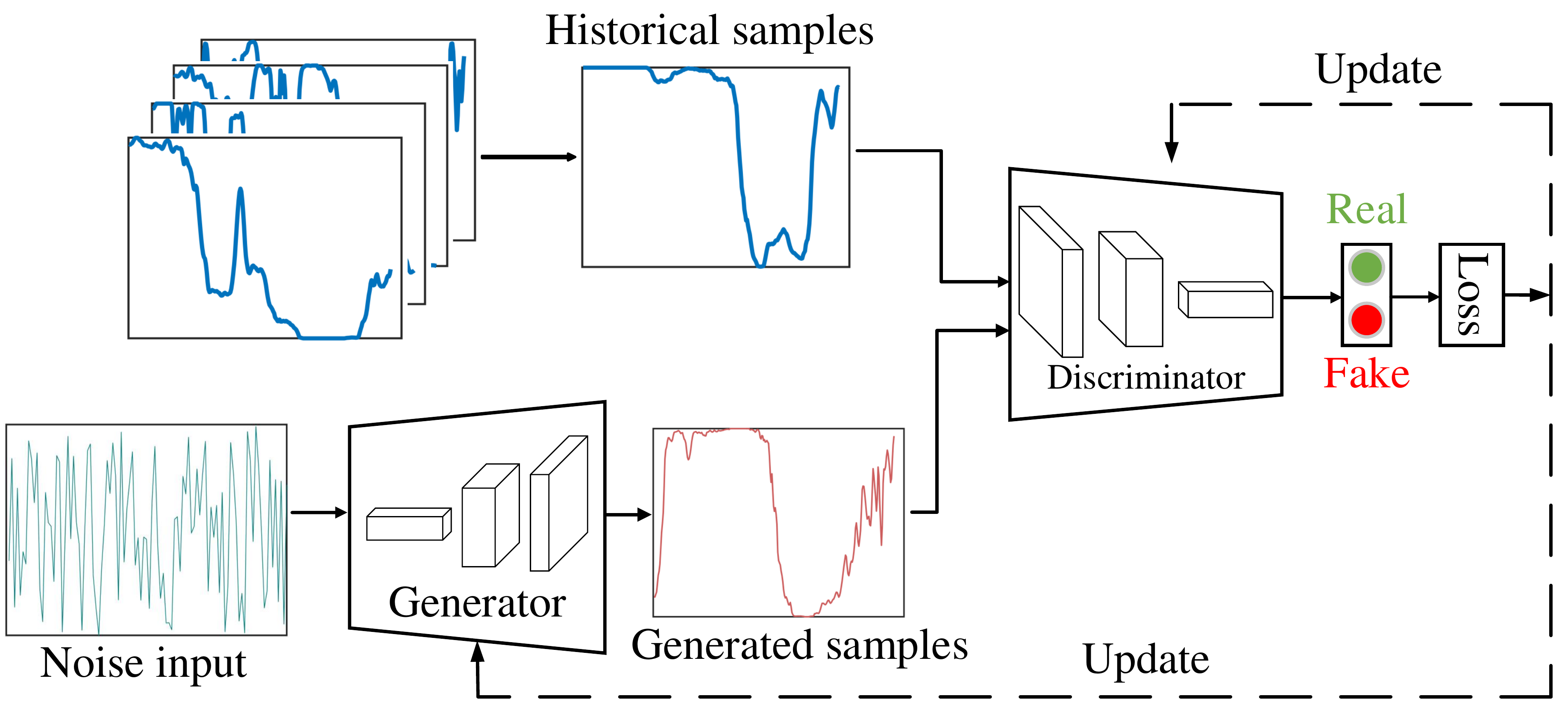}	
	\caption{Basic architecture of GANs.}	
	\label{fig_architecture}
\end{figure}

GANs is an unsupervised generative model based on game theory proposed by Goodfellow \cite{goodfellow2014generative}. Fig. \ref{fig_architecture} shows the basic architecture of GANs, which includes two deep neural networks: discriminator ($D$) and generator ($G$). The generator network maps random noises to generated samples by learning the potential distribution of historical data, and the discriminator judges whether the input data is the real historical data or the generated data as much as possible.

For convenience, $\bm{x}$ represents the real historical data; $P_{d}$ denotes the real historical data distribution, which is an unknown and difficult to model distribution. This study uses $\bm{z}\sim P_Z$ as the random noise vector, where $P_Z$ is a known distribution that is easy to sample (e.g., Gaussian distribution). Let differentiable functions $G(\cdot ;\theta_g)$ and $D(\cdot ;\theta_d)$ be the generator function and the discriminator function, where $\theta_g$ and $\theta_d$ are the weight parameter of these two networks.

During training, a batch of random noise vectors $\bm{z}\sim P_Z$ are fed into the generator, which is mapped from the noise space to the data space $G(\bm{z};\theta_g)$ obeying the distribution $P_g$. Meanwhile, a batch of samples that come from real data $\bm{x}$ or generated data $G(\bm{z})$ are sent to the discriminator that is usually a binary classifier whose goal is to identify which the data comes from. According to different data sources, the outputs of the discriminator network can be expressed as
\begin{equation}
    \label{eq.PD}
    \left\{
      \begin{aligned}
        p_{real}&=D(\bm{x}) \\
        p_{fake}&=D(G(\bm{z})) 
      \end{aligned}
  \right.
\end{equation}

After clarifying the goals of the generator and the discriminator, we need to respectively define a loss function for them to update their deep neural networks. For a given discriminator, the generator wants to increase the probability output of the discriminator to the generated samples $p_{fake}$ since a larger discriminator output means the sample is more realistic. For a given generator, the discriminator seeks to the maximization of $p_{real}$ while minimizing $p_{fake}$. In other words, a small generator loss value $L_G$ suggests that the manifolds of the generated data and real data are extremely close, while a small value of $L_D$ demonstrates that the discriminator has a strong ability to distinguish which distribution the data comes from. So, the loss functions of the two neural networks can be defined as
\begin{equation}
    \label{eq_Gloss}
    L_G = \mathbb{E}_{\bm{z}\sim P_Z}[\log (1-D(G(\bm{z})))]
\end{equation}
\begin{equation}
    \label{eq_Dloss}
    L_D = -\mathbb{E}_{\bm{x}\sim P_{d}}[\log D(\bm{x})]-\mathbb{E}_{\bm{z}\sim P_{d}}[\log (1-D(G(\bm{z})))]
\end{equation}
where $\mathbb{E}$ denotes the expected value of the batch samples.

In summary, we need to construct a game relationship between generator and discriminator so that the two networks can be trained simultaneously. It is necessary to build a value function that can combine  (\ref{eq_Gloss}) and (\ref{eq_Dloss}). The mini-max game model with value function $V_{\text{GANs}}(G, D)$ is given by \cite{goodfellow2014generative}:

\begin{equation}
    \begin{aligned}
        \label{eq.GANs}
        \min _{G} \max _{D} V_{\text{GANs}}(G,D)=&\mathbb{E}_{\bm{x} \sim P_{d}}[\log D(\bm{x})]\\&+\mathbb{E}_{\bm{z} \sim P_{z}}[\log (1-D(G(\bm{z})))]
    \end{aligned}
\end{equation}

\subsection{Federated Learning}
Traditional centralized methods need to collect training data into a central data center and perform centralized training, which will cause data security issues and huge communication overheads. To cope with these problems, we introduced an emerging and powerful distributed machine learning technology---federated learning.

Federated learning is initially proposed by Google research team to establish a keyboard prediction model of mobile phones~\cite{mcmahan2017communication}. Suppose there are $N$ clients, i.e. participating edge devices, $\{\mathcal{C}_1,\mathcal{C}_2,\ldots,\mathcal{C}_N \}$ that want to train a learning model based on their own datasets $\{\mathcal{D}_1,\mathcal{D}_2,\ldots,\mathcal{D}_N \}$. A traditional centralized method aims to construct a large-scale dataset $\mathcal{D}=\mathcal{D}_1 \cup\cdots \mathcal{D}_N$ and train a model $\mathcal{M}_\text{SUM}$ by collecting all the data together in a central workstation. For the centralized method, any client will expose his data to the workstation or other clients. Federated learning coordinates clients to train a global model $\mathcal{M}_\text{FED}$ deployed on a central server, rather than collecting data from all clients. In addition, assuming that $\mathcal{V}_{\text{SUM}}$ and $\mathcal{V}_{\text{FED}}$ are the performance metrics of the centralized model $\mathcal{M}_\text{SUM}$ and federated model $\mathcal{M}_\text{FED}$ respectively. Formally, let $\delta$ be a non-negative real number, if
\begin{equation}\label{eq.federated}
    \left| \mathcal{V}_{\text{SUM}} -\mathcal{V}_{\text{FED}}\right|<\delta
\end{equation}
we say the federated learning algorithm has $\delta$-accuracy loss. Eq.(\ref{eq.federated}) suggests the fact that if federated learning is used to build a learning model on distributed data sources, the performance of this model is similar to that of a centralized model obtained by training all the data together~\cite{yang2019federated}.

\begin{figure}
    \centering
    \includegraphics[width=3.4in]{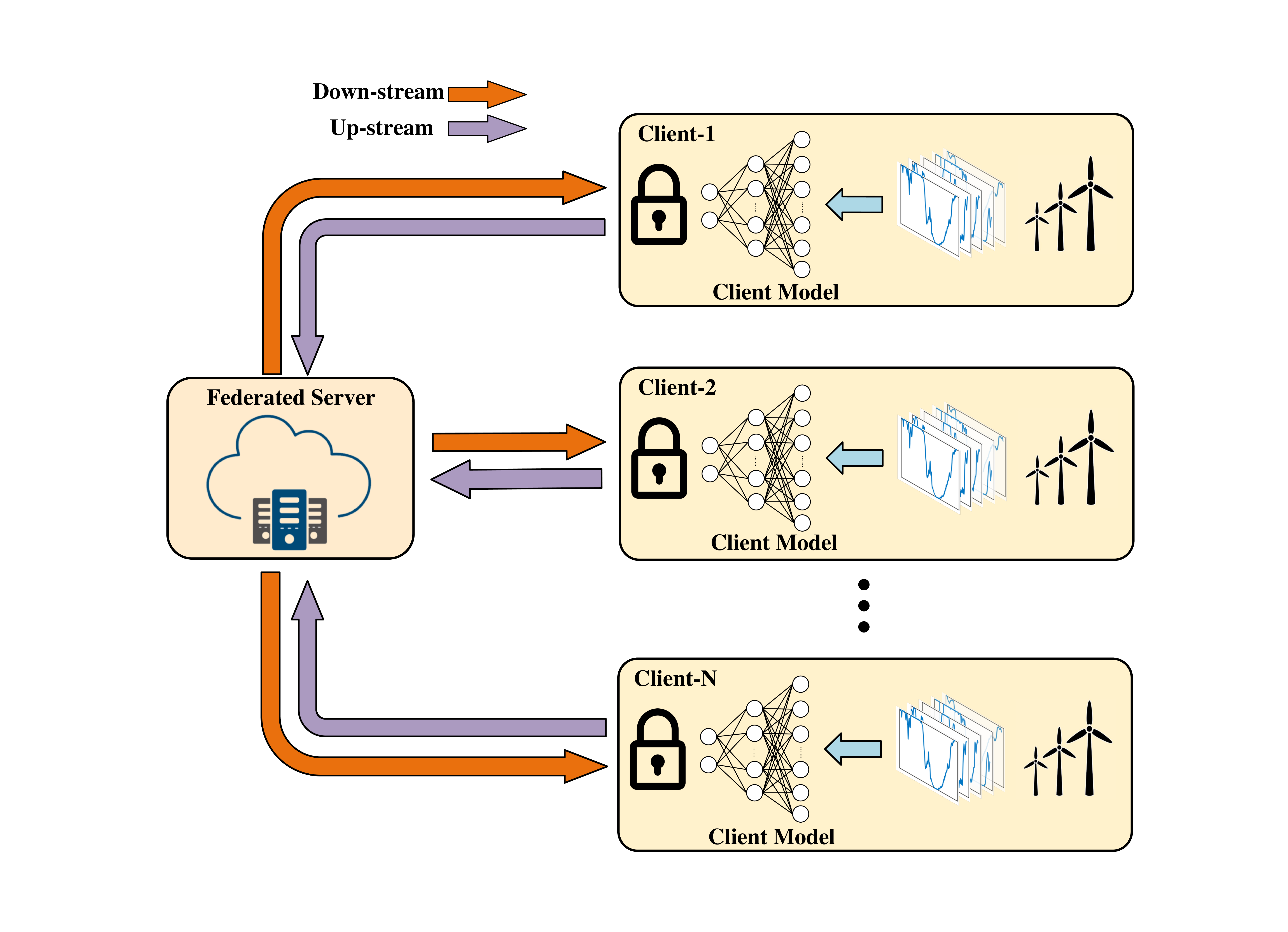}
    \caption{The basic structure of server-client in federated learning.}
    \label{fig:federated_learning}
\end{figure}

The client-server federated learning architecture is shown in Fig.~\ref{fig:federated_learning}. Firstly, the central server sends the initialized global model to each client; while the clients use the data to train their own models locally, and then send the models' parameters to the central server through the communication system. Next, the server aggregates the model parameters uploaded by each client and updates the weights parameters of the shared global model. And then, the updated weight parameters are distributed to each client. The above interaction learning process is repeated until a prespecified training termination criterion is satisfied.

\section{Fed-LSGAN for Renewable Scenario generation}\label{sec.fedgan}

\subsection{Problem Formulation}

Due to the inherent randomness and intermittency of renewable resources, wind and solar powers can be modeled as a stochastic sequence though scenario generation methods, which is condictive to the analysis of the uncertainty of renewable energies. In order to solve the problem that renewable energy power sites belonging to different owners like ISOs are unwilling to share data due to business competition factors and data privacy, this work focuses on scenario generation based on federated learning technology. We refer to renewable energy power sites as clients, and use a transmission or distribution system operator as the central server to generate time series scenarios of renewable energy outputs. Furthermore, because there are multiple renewable generation sites which are relatively close in geographical location, the spatial correlation between sites must also be considered during scenario generation, besides the time correlation in each site.

For a group of $N_{r}$ renewable energy power sites $\mathcal{C}_i, i=1,2,\ldots, N_r$, the historical power generation data is utilized as a training set. Let the historical data of each site be $\mathcal{D}_i=\{\bm{x}_1, \bm{x}_2,\ldots,\bm{x}_T\},i=1,2,\ldots, N_r$, and use it as the input data of the local model. Each site saves the data locally and uses edge devices with sufficient energy supply deployed in renewable energy sites to train local models. During training, each edge device communicates with the server via the federated learning mechanism. Our goal is to obtain a shared global model that generates renewable scenarios with spatial characteristics and local models that yield the scenarios reflecting  temporal correlation of a client. 

\subsection{Global LSGANs Model}
\subsubsection{Least Square Generative Adversarial Networks}

In the traditional GANs, when the generator is fixed, the optimal discriminator is as follows:
\begin{equation}\label{eq.D_of_GANs}
  D^*_{G,\text{GANs}}(\bm{x})=\frac{P_{d}(\bm{x})}{P_{d}(\bm{x})+P_g(\bm{x})}
\end{equation}

For the sake of achieving the Nash equilibrium of the two networks, the generator needs to be minimized. Substituting (\ref{eq.D_of_GANs}) into (\ref{eq.GANs}), we get the conclusion that minimizing  (\ref{eq.GANs}) yields the minimization of the Jensen-Shannon divergence (JSD) between the two distributions:
\begin{equation}
    C_{\text{GANs}}(G)=V\left(D_{D,\text{GANs}}^{*}, G\right)=2\text{JSD}\left(P_{d} \| P_{g}\right)-\log (4)
\end{equation}

The JSD is always non-negative, which is equal to zero only when the two distributions are completely equal, and $C_{\text{GANs}}^*(G)=-\log 4$ is the global minimum of $C_{\text{GANs}}(G)$ when $P_g=P_{d}$. However, the two distributions will not overlap, or the overlap can be ignored when the $P_{d}$ and $P_g$ are high-dimensional distributions. At this time, if the two distributions cannot overlap in all dimensions, the JSD always remains very close to $\log 2$. Consequently, the Jensen-Shannon distance saturates, and the discriminator has zero loss. Eventually, due to the use of cross-entropy loss function, the traditional GANs suffers from the issues such as mode collapse, vanishing gradient and the instability in training.

To resolve this problem,  reference \cite{mao2019effectiveness} proposes the LSGANs, which use $a$-$b$ encoding and the least squares loss function. The objective function of LSGANs is 
\begin{equation}\label{eq.LSGANs}
    \begin{aligned}
    \min _{D} V_{\text{LSGANs}}(D) &=\frac{1}{2} \mathbb{E}_{\bm{x} \sim P_{d}}\left[(D(\bm{x})-b)^{2}\right]+\\ &\frac{1}{2} \mathbb{E}_{\bm{z}  \sim P_{Z}}\left[(D(G(\bm{z} ))-a)^{2}\right] \\
    \min _{G} V_{\text{LSGANs}}(G) &=\frac{1}{2} \mathbb{E}_{\bm{z}  \sim P_{Z}}\left[(D(G(\bm{z}))-c)^{2}\right]
    \end{aligned}
\end{equation}
where $c$ is the value which the generator wants the discriminator to trust for generated data; $a$ and $b$ are respectively labels for generated data and real data. 

For a fixed generator $G$, the optimal discriminator $D$ is
\begin{equation}
  D_{G,\text{LSGANs}}^{*}(\bm{x})=\frac{b P_{d}(\bm{x})+a P_{g}(\bm{x})}{P_{d}(\bm{x})+P_{g}(\bm{x})}
\end{equation}

Next, Eq.(\ref{eq.LSGANs}) can be reformulated as
\begin{equation}
  C_{\text{LSGANs}}(G)=\int_{\mathcal{X}} \frac{\left((b-c) P_{d}(\bm{x})+(a-c) P_{g}(\bm{x})\right)^{2}}{P_{d}(\bm{x})+P_{g}(\bm{x})} \mathrm{d} x
\end{equation}

We assume that $b-c=1$ and $b-a=2$, then 
\begin{equation}
  2 C_{\text{LSGANs}}(G)=\chi_{\mathrm{Pearson}}^{2}\left(P_{d}+P_{g} \| 2 P_{g}\right)\
\end{equation}
where $\chi_{Pearson}^2$ is the Pearson $\chi^2$ divergence. If $b-c = 1$ and $b-a=2$ are satisfied,  (\ref{eq.LSGANs}) is equivalent to minimize the Pearson $\chi^2$ divergence between $P_{d}+P_g$ and $2P_g$. In this case, as long as the two distributions are not completely overlap, the loss will not be zero~\cite{nowozin2016f}. In this work,  we set $a = 0$, $b = c = 1$. LSGANs leverage the least squares loss function for discriminator and generator, which can provide more gradients through penalizing samples far from the decision boundary, relieving the problem of vanishing gradients. Meanwhile, this loss function is beneficial for the generation of higher quality samples since it penalizes the fake samples and ``pull'' them toward the decision boundary.

\subsubsection{LSGANs Configuration}
According to the characteristics of renewable generation data, the deep neural networks of generator and discriminator in global LSGANs under the federated learning framework are designed in this paper. 
\begin{figure}
    \centering
    \includegraphics[width=3.4in]{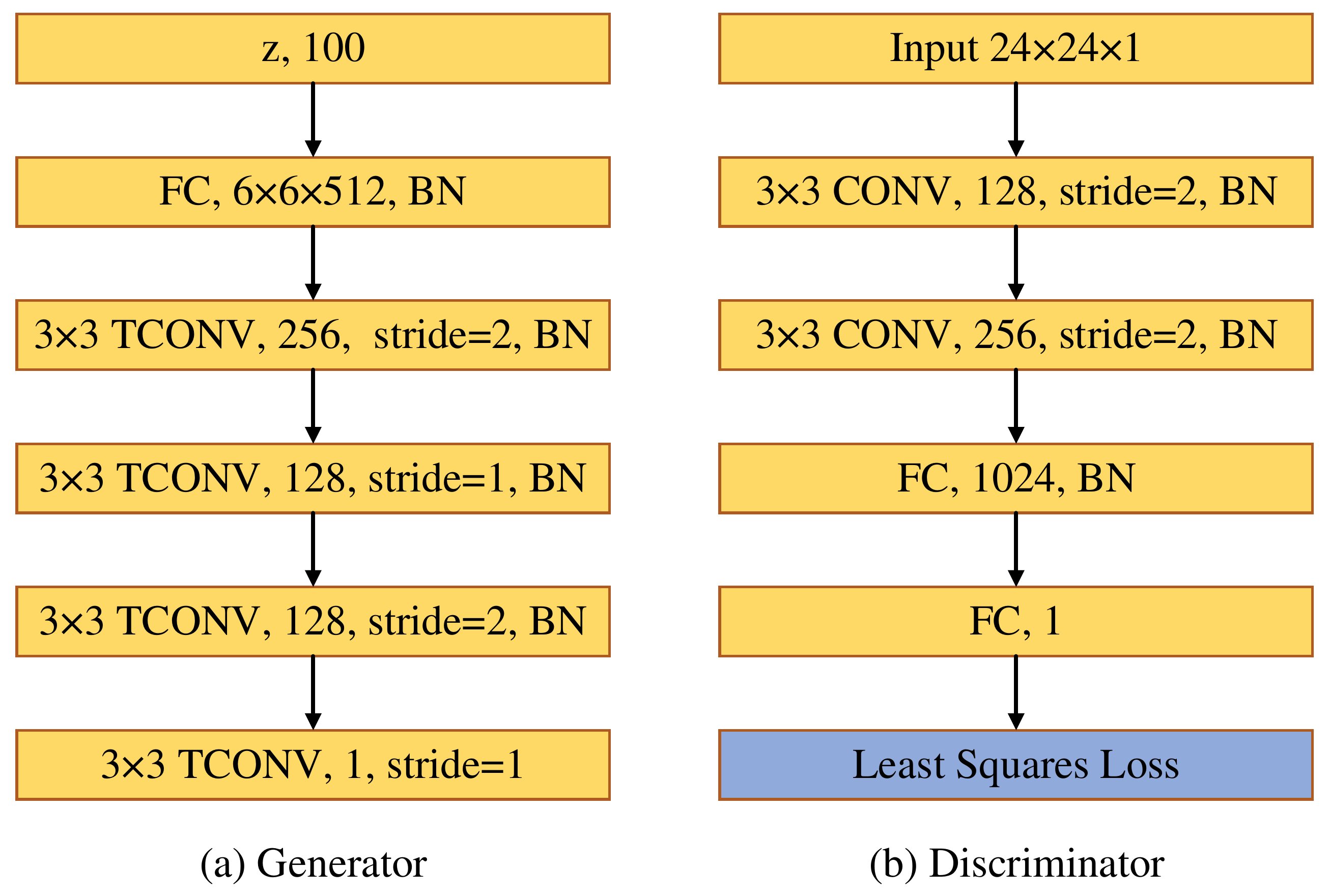}
    \caption{The global LSGANs model structure. ``$B\times B$'' indicates that the kernel size is $B$. ``CONV, TCONV, FC'' respectively denote the convolutional layer, the transposed convolutional layer and the fully-connected layer. BN indicates that this layer uses the batch normalization.}
    \label{fig.model_structure}
\end{figure}

Fig. \ref{fig.model_structure} shows that the details of generator and discriminator network structures. These networks contain fully connected neural network layer and convolution neural network layer. With the exception of the output layer, we adopt the ReLU and LeakyReLU activation functions to the generator and discriminator, respectively. Our model is trained using the Adam optimizer, where the parameter $\beta_1$ is set to 0.5. At the same time, the batch normalization is introduced into different layers (expect the input and output layers) to speed up the training through reducing internal covariate shift~\cite{ioffe2015batch}. All weights of neurons in neural network are initialized by central normal distribution with the standard deviation 0.02.

\subsection{Fed-LSGAN Framework}
\begin{figure}
    \centering
    \includegraphics[width=3.3in]{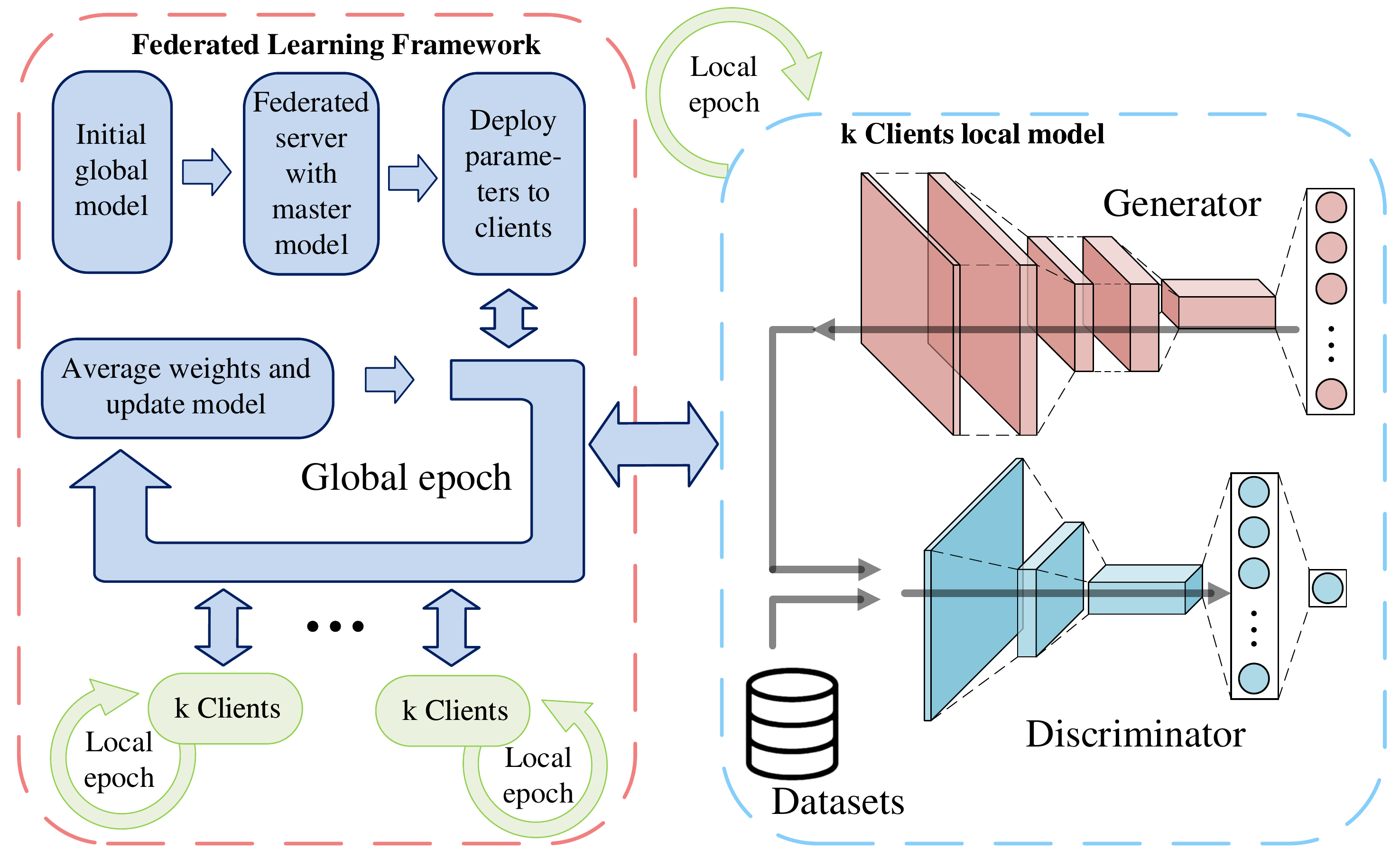}
    \caption{Training process of client-server architecture in federated learning. }
    \label{fig.fed_gan}
\end{figure}

 Based on the above theoretical analysis, we propose the Fed-LSGAN to generate renewable scenarios, where \emph{FederatedAveraging} (\emph{FedAvg}) algorithm is adopted in federated settings. The Fed-LSGAN yields a shared global model through aggregating separated local models without sacrificing data privacy. The flowchart of our method is shown in Fig.~\ref{fig.fed_gan}.

To facilitate analysis, we assume that all clients agree in advance to train a deep neural network model with an identical initial global framework, and use the same optimization algorithm and training hyperparameters (e.g. learning rate, minibatch size) in the proposed Fed-LSGAN. In addition, the computing powers of the clients are assumed to be similar; the local epoch of each client is taken as 1; and the data packet loss during communications is not considered.

To sum up, the details of the Fed-LSGAN are presented in Algorithm~\ref{alg:the_alg}. The Fed-LSGAN is mainly controlled by the following parameters: $E$, the fraction of clients performing computation during training; $K$, synchronization interval. In this work, we set $K=100$ and $E=100\%$. Note that the central server does not train its  generator and discriminator by itself, and the parameters of the global model are only obtained through aggregating those of the local models.

\begin{algorithm}[!ht] 
    \small
    \caption{Federated Least Squares Generative Adversarial Networks (Fed-LSGAN)} 
    \label{alg:the_alg}
	\begin{algorithmic}[1] 
    \REQUIRE Learning rate $\alpha$; Adam hyperparameters $\beta_1$, $\beta_2$; Local minibatch size $m$; Global epoch $W$; Synchronization interval $K$, Client ratio $E$.
    \REQUIRE A global LSGANs model with parameters $(\theta_g^\mathcal{S}, \theta_d^\mathcal{S})$ for Discriminator and Generator on central server $\mathcal{S}$; local LSGANs models with parameters $\{\theta_g,\theta_d\}_{i=1}^{N_r}$ on $N_r$ clients $\{ \mathcal{C}\}_{i=1}^{N_r}$.
    \FOR{global epoch $w=1,2,\ldots, W$}
    \STATE Randomly select $N_e$ clients from all clients $ \{\mathcal{C}\}$ using $E$
    \FOR{each selected client $e$ \textbf{in parallel}} 
    \STATE Update discriminator $\theta_d^e$ of client $e$
    \STATE $\bullet$ Sample batch noise samples $\{z\}_{i=1}^m\sim P_{Z}$.
    \STATE $\bullet$ Sample batch samples $\{x\}_{i=1}^m$ from training set.
    \STATE $\bullet$ Update parameters for discriminator network
    \begin{equation*}
     \left\{\begin{array}{l} g_{\theta_d}  \leftarrow  \nabla_{\theta_d}\frac{1}{m}\sum_{i=1}^m[\frac{1}{2}(D(x^{(i)})-1)^2+
        \frac{1}{2}(D(G(z^{(i)})))^2]
        \\ \theta_d  \leftarrow\theta_d-\alpha\cdot \text{Adam}(\theta_d,g_{\theta_d},\beta_1,\beta_2)\end{array} \right.
    \end{equation*}
    \STATE Update generator $\theta_g^e$ of client $e$
    \STATE $\bullet$ Sample batch noise samples $\{z\}_{i=1}^m\sim P_{Z}$.
    \STATE $\bullet$ Update parameters for generator network
    \begin{equation*}
     \left\{\begin{array}{l}g_{\theta_g}\leftarrow\nabla_{\theta_g}\frac{1}{m}\sum_{i=1}^m\frac{1}{2}(D(G(z^{(i)}))-1)^2\\\theta_g\leftarrow\theta_g-\alpha\cdot \text{Adam}(\theta_g,g_{\theta_g},\beta_1,\beta_2)\end{array} \right.
    \end{equation*}
    \ENDFOR
    \IF {$w \mod K = 0$} 
        \STATE All selected clients send parameters to server, and the server aggregate $\theta^\mathcal{S}$ by averaging
        \begin{equation*}
            \theta_g^\mathcal{S}\leftarrow\frac{1}{N_e}\sum_{e=1}^{N_e}\theta_g^e,\quad \theta_d^\mathcal{S}\leftarrow\frac{1}{N_e}\sum_{e=1}^{N_e}\theta_d^e
        \end{equation*}
        \STATE The server send back parameters and clients update local parameters
        \begin{equation*}
            \{\theta_g,\theta_d\}_{i=1}^{N_r}\leftarrow(\theta_g^\mathcal{S}, \theta_d^\mathcal{S})
        \end{equation*}
    \ENDIF
    \ENDFOR
  \end{algorithmic}
\end{algorithm}

\section{Case study}\label{sec.case_study}
This section first analyzes temporal characteristics of client models and spatial characteristics of global model by visually comparing with generated data and real data. The effectiveness and superiority of our method are examined based on qualitative  and  quantitative  evaluation  results. Besides, we design an experiment to examine the robustness of the Fed-LSGAN with different federated  settings. In this study, the Fed-LSGAN is implemented with PyTorch, which is a library for Python programs that facilitates building deep learning projects.

\subsection{Data Description}
The used wind and solar power data set are collected from the integrated dataset from National Renewable Energy Laboratory (NREL) \cite{draxl2015wind}. The dataset provides a large amount of renewable energy generator data with a sampling interval of 5 minutes. In this paper, 32 wind sites and 12 solar sites in Washington State, USA are adopted as the client's data set. In this study, each client saves the data locally, and takes 80\% of the data set as the training set and 20\% as the test set. As the geographical locations of the selected sites are close to each other, there is a certain correlation between the output of each site. 

\subsection{Temporal Correlation Analysis of Client Model}\label{sec.single_site}
Without loss of generality, we take the scenarios generated by a client as an example to analyze the temporal correlation of a client model. In the proposed Fed-LSGAN, each client trains its local model by using the data stored in network edge devices. As the NREL dataset consists of 288 points per day, 576 points in two days are reshaped into $24\times 24$ matrix as the input of each client's local model. 
\begin{figure}
    \centering
    \includegraphics[width=3.4in]{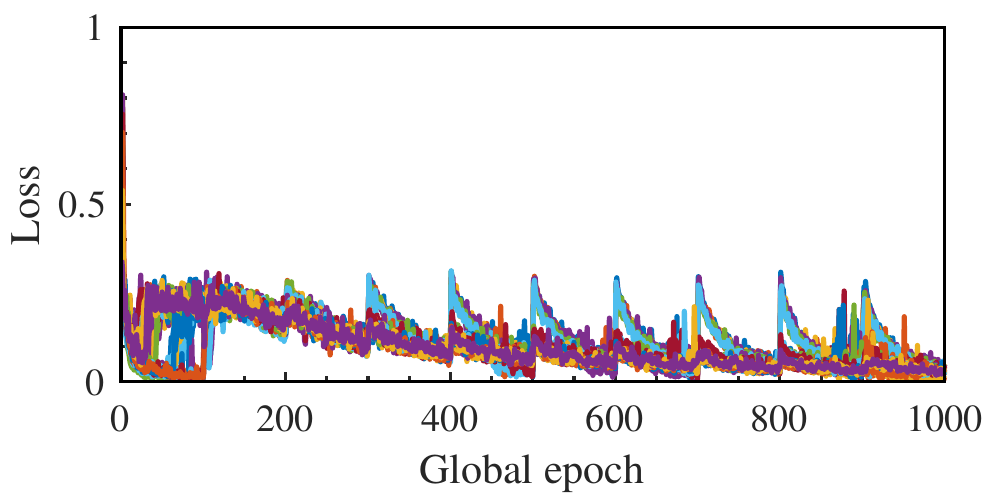}
    \caption{The discriminator loss value of each client during the training process.}
    \label{fig.Loss}
\end{figure}

The discriminator loss value of each client during training is shown in Fig. \ref{fig.Loss}. It can be seen from the figure that the loss value drops rapidly at the beginning of training, and then rises vertically every 100 global epochs afterwards. This is because after the client communicates with the central server, the server distributes the aggregated parameters to each client, which causes the client loss value to increase sharply. And eventually, the discriminator loss value of each client will approach 0, which indicates that the distribution of the generated scenarios is sufficiently close to the real data distribution. So far, the global and local models are obtained.

\subsubsection{Time Series Analysis}

\begin{figure*}[!t]
    \centering
    \subfloat[wind]{\includegraphics[width=6in]{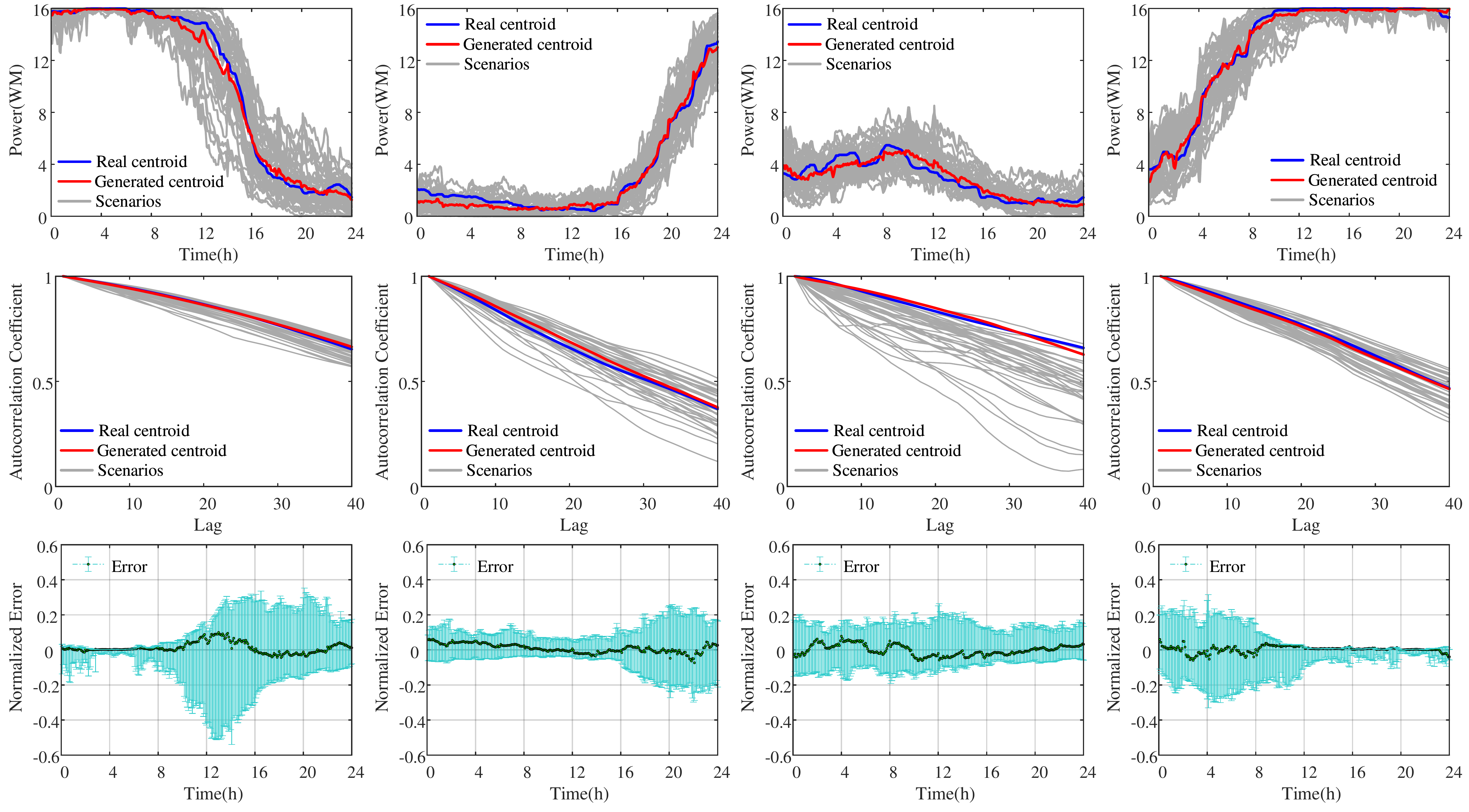}
    \label{fig.wind_samples}}
    \hfil
    \subfloat[solar]{\includegraphics[width=6in]{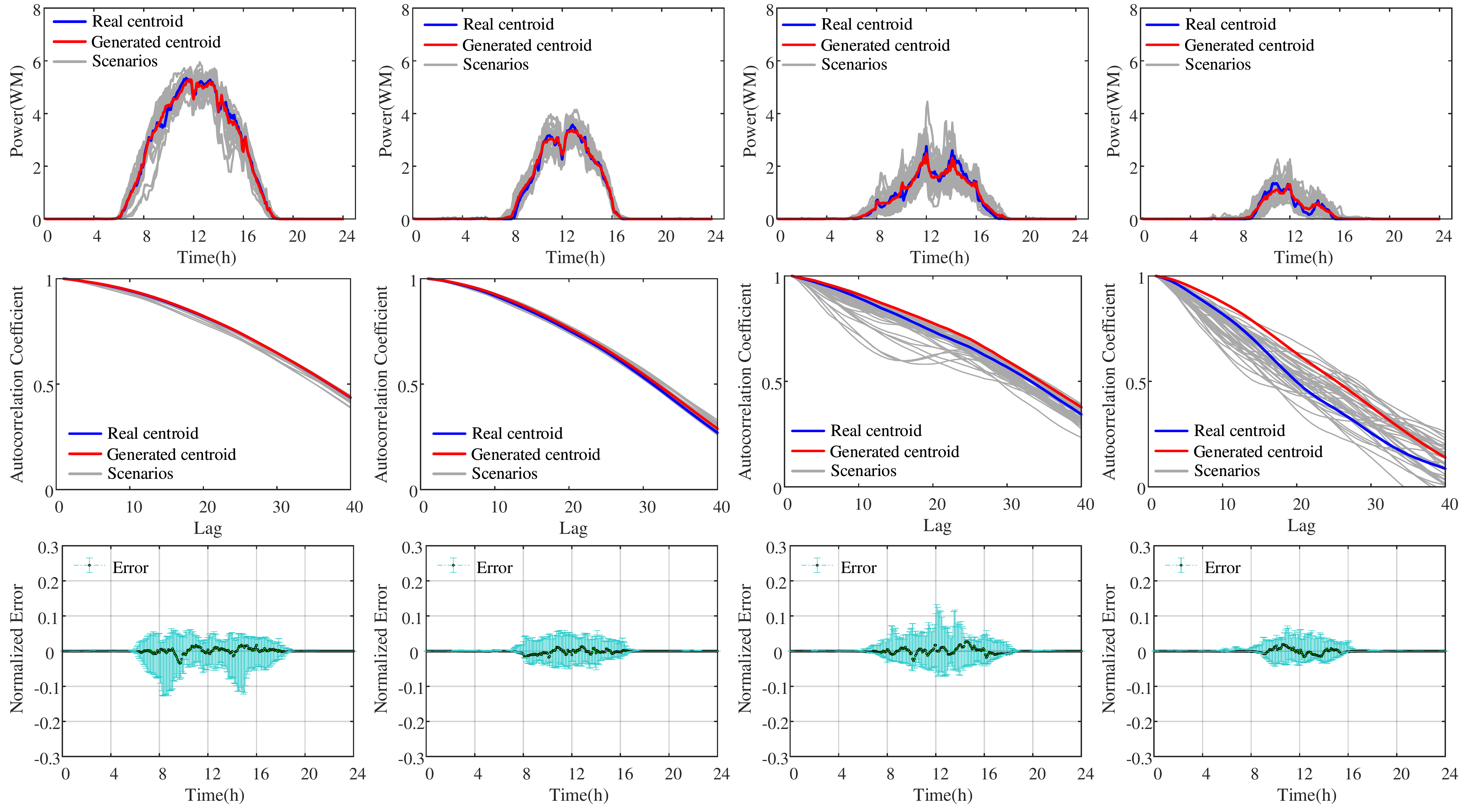}
    \label{fig.solar_samples}}
    \caption{The generated wind and solar power scenario. The top row of (a) and (b) is the comparison of the generated scenario and its centroid with the centroid of the test set. The middle row is the autocorrelation coefficient of the scenario. The bottom row is the normalized error between the generated scenario and the centroid of the test set.}
    \label{fig.samples}
\end{figure*}

The k-means clustering algorithm is adopted on the generated scenarios and test sets to obtain renewable energy scenario clusters with different random characteristics. For clarity, some scenario clusters are shown in Fig. \ref{fig.samples} to illustrate the generated scenarios. These figures show that our method is capable of reproducing the power characteristics of wind (e.g. ramping characteristics, power fluctuations) and solar (e.g. day-night changes). By comparing the test set and the generated scenarios, we find that the Fed-LSGAN can capture the significant characteristics of renewable power generations under the premise of protecting the data privacy of renewable energy power plants. Based on this phenomena, we can safely draw a conclusion that the Fed-LSGAN can generate new samples with good diversity that capture the underlying intrinsic manifolds of the real data, rather than simply simulating the training data. 

In the bottom rows of Figs.~\ref{fig.wind_samples} and~\ref{fig.solar_samples}, the normalized errors of the generated scenarios are visualized, where the upper and lower limits of an error bar are the maximum and minimum normalized errors of each scenario. One can see that for wind power scenarios, the normalized error of the centroid of each generated scenarios relative to that of the real data cluster does not exceed 0.1, and the error limits of each scenario are less than 0.5; while regarding solar scenarios, the normalized errors are much smaller than those of wind power scenarios.

\subsubsection{Correlation Analysis}

The correlation of time series plays a vital role in the operation and planning of power systems, since it can reflect the practical operating conditions of renewable energy in terms of temporal characteristics, which can avoid underestimating operating costs~\cite{villanueva2011simulation}. Therefore, it is necessary to examine the effectiveness of the generated scenarios from the perspective of correlation analysis. The autocorrelation coefficient measures the degree of correlation of time series in two different periods. Vividly speaking, it measures the impact of past behaviors on the present. We calculate the autocorrelation coefficients $R(\tau)$ of the samples by

\begin{equation}\label{eq.correlation}
    R(\tau) = \frac{\mathbb{E}[(S_t-\mu)(S_{t+\tau}-\mu)]}{\sigma^2}
\end{equation}
where $S$ is a random time series; $\mu$ and $\sigma$ denote the mean and variance of $S$, respectively; and $\tau$ is the time lag. 

We show the correlation coefficients of real and generated samples in the middle rows of Figs.~\ref{fig.wind_samples} and~\ref{fig.solar_samples}. It can be known that regarding the temporal correlation characteristics, the generated scenarios almost perfectly reproduce the characteristics of the real data while maintaining its diversity, and fully represent the practical operating conditions of the wind/solar sites.

\subsubsection{Statistical Characteristic Analysis}
\begin{figure}
    \centering
    \subfloat[wind]{\includegraphics[width=3.4in]{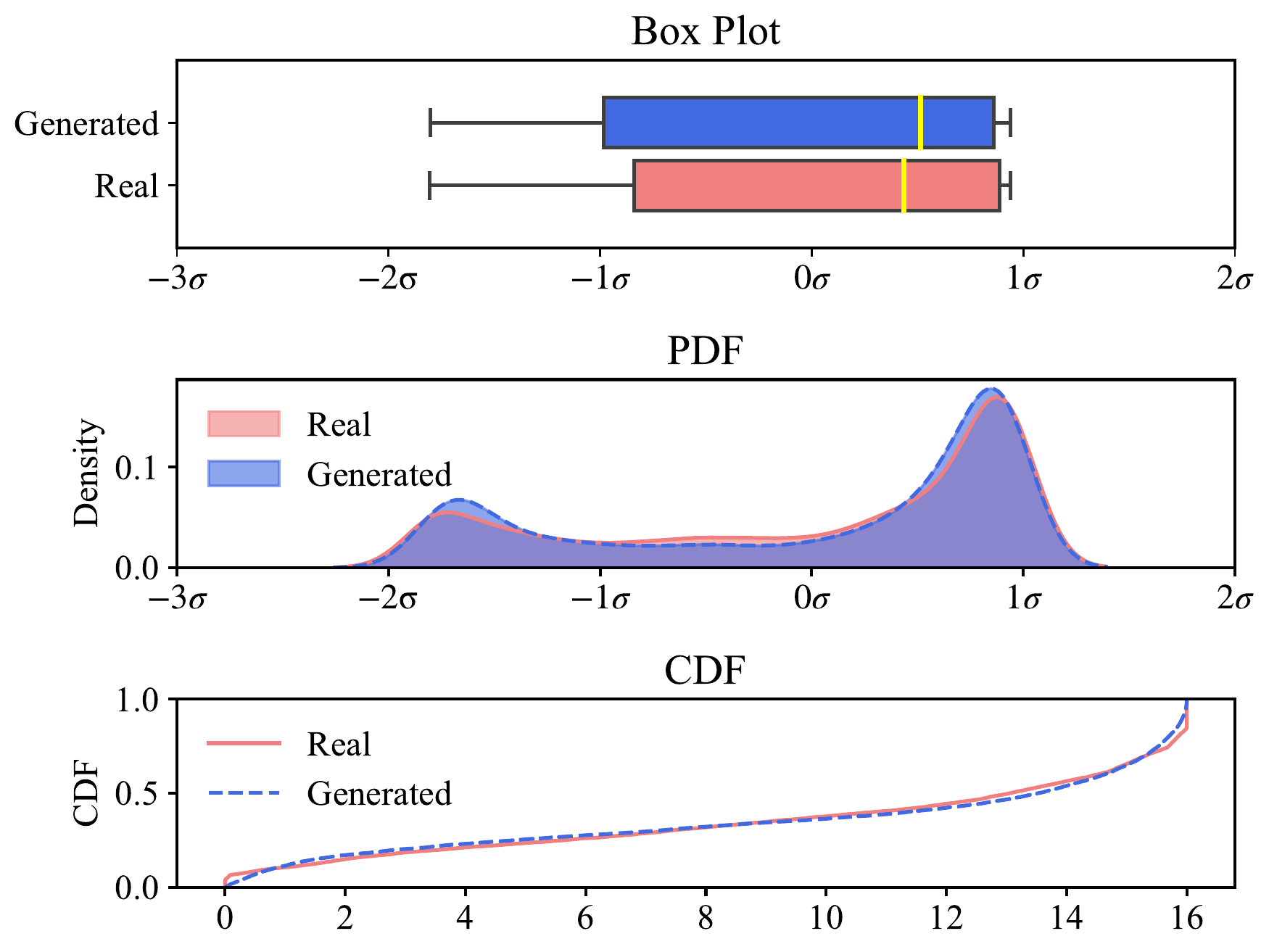}
    \label{fig.wind_statis}}
    \hfil
    \subfloat[solar]{\includegraphics[width=3.4in]{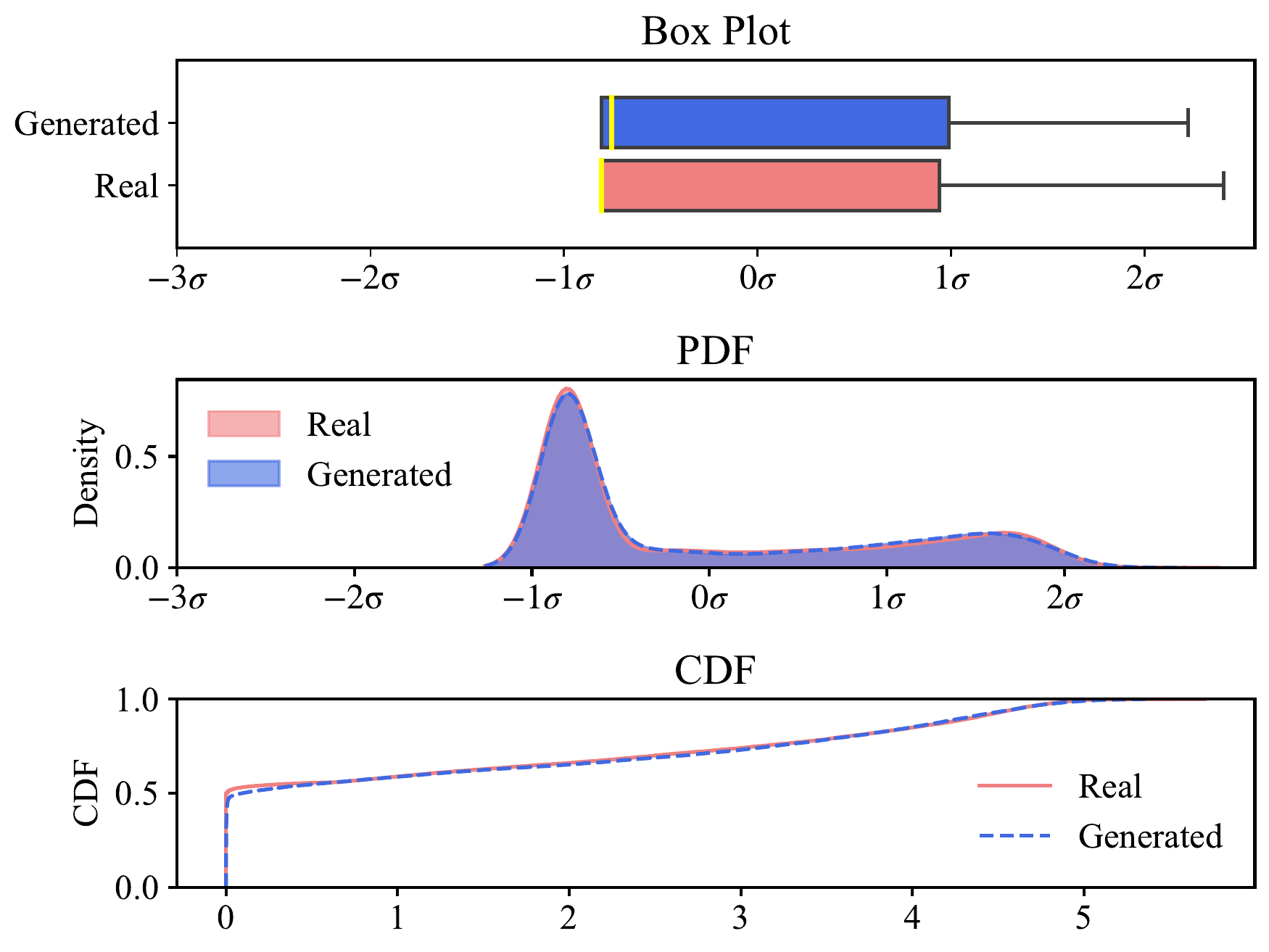}
    \label{fig.solar_statis}}
    \caption{The statistical characteristics of wind and solar power are analyzed from three aspects: box plot(top), PDF(middle) and CDF(bottom). The yellow line in the box plot represents the median.}
    \label{fig.statis}
\end{figure}

Besides the above analysis, the long-term characteristics of the generated scenarios are analyzed. To this end, we demonstrate the statistical characteristics of the generated scenarios in detail from three aspects: box plot, probability density function (PDF), and cumulative distribution function (CDF). As shown in Fig.~\ref{fig.statis}, the Fed-LSGAN captures the statistical details of the renewable generation outputs, which suggests our approach manages to fully learn the distribution of real data.

For wind scenarios, as shown in Fig.~\ref{fig.wind_statis},  the PDFs of real samples and generated samples are almost the same, and both of them show negatively skewed; the median and the interquartile ranges are roughly equivalent from the box plot. The same conclusion can be drawn from the CDF curves, that is, the probability of wind power outputs in the interval $[13,16]$ MW reaches 50\%. For solar scenarios, the generated samples also have similar statistical characteristics to the real samples from Fig.\ref{fig.solar_statis}. However, the statistical characteristic of solar powers are obviously different from that of wind powers, because solar power generations can only work during the daytime. The CDF curves of solar power illustrate even the probability that the photovoltaic output is $0$ MW reaches 50\%. Therefore, from the perspective of statistical analysis, a conclusion can be drawn that the Fed-LSGAN manages to generate high-quality scenarios by using federated deep generative learning.

\subsubsection{Comparison with Other Methods}
\begin{figure}
    \centering
    \subfloat[wind]{\includegraphics[width=3.4in]{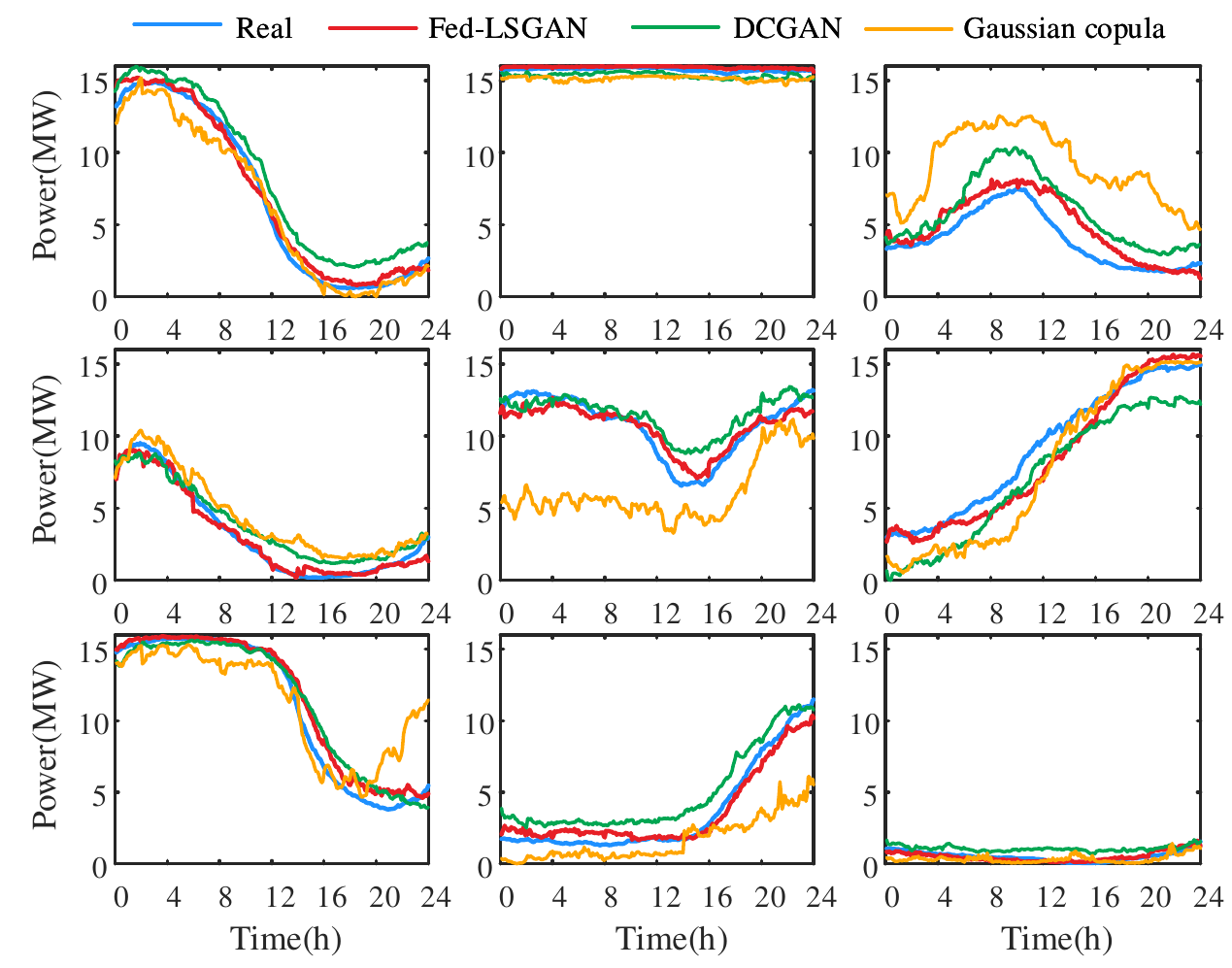}
    \label{fig.wind_kmeans}}
    \hfil
    \subfloat[solar]{\includegraphics[width=3.4in]{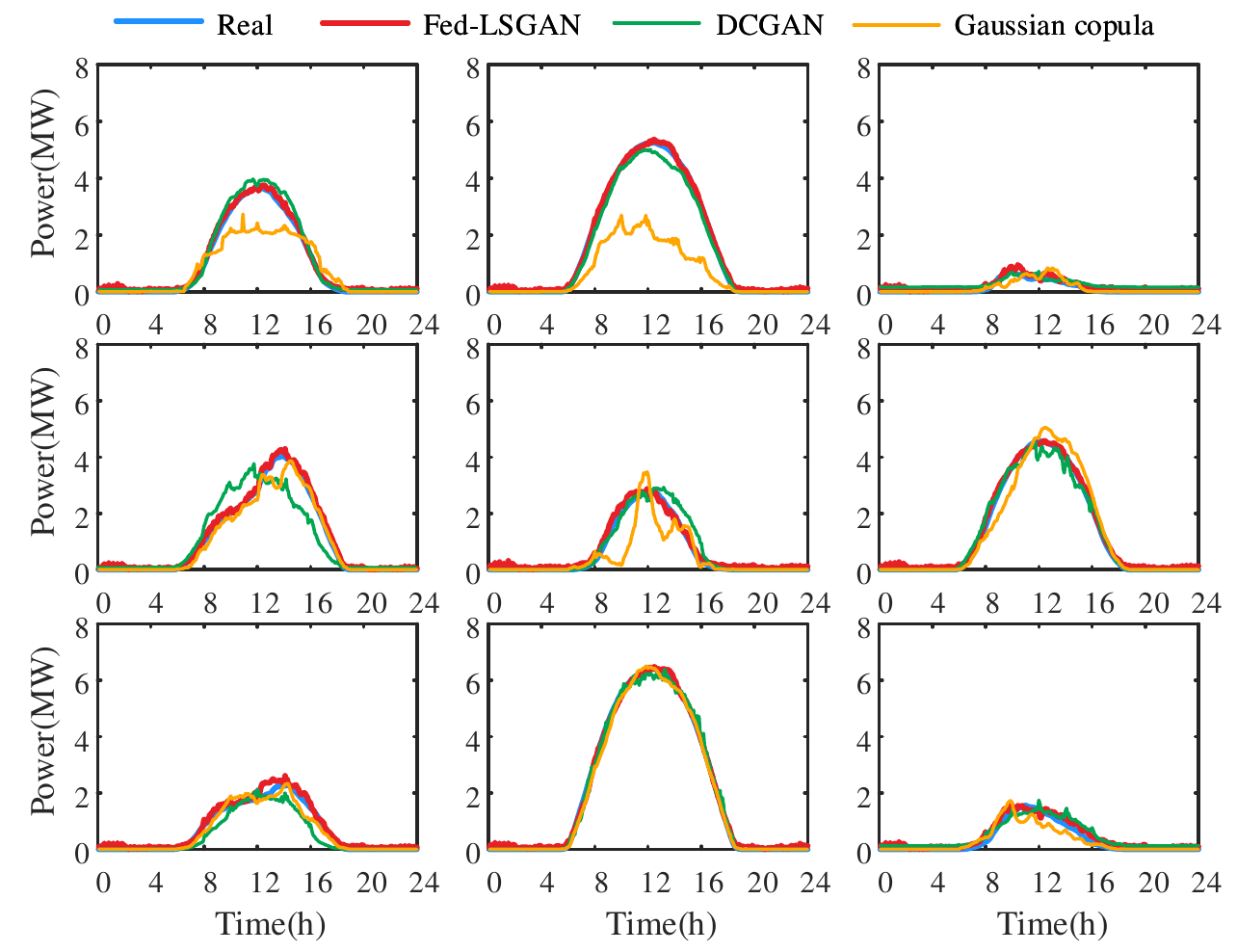}
    \label{fig.solar_kmeans}}
    \caption{Clustering results of scenarios generated by different methods.}
    \label{fig.kmeans}
\end{figure}

In order to properly evaluate the performance of the proposed Fed-LSGAN, we compare it with typical centralized scenario generation methods such as deep convolutional GAN (DCGAN) and Gaussian copula. In this experiment, the k-means algorithm is adopted for clustering the real data and the generated data into 9 clusters, respectively; and the centroids of these clusters are shown in Fig.~\ref{fig.kmeans}. This figure shows that the Fed-LSGAN is able to generate high-quality scenarios that are sufficiently close to the actual data. Consequently, the following conclusions can be made: (1) In terms of scenario generation quality, the  Fed-LSGAN significantly outperforms the model-based Gaussian-copula method and is better than the centralized DCGAN; (2) Most importantly, the Fed-LSGAN is capable of protecting data privacy and reducing communication expenses by using the federated learning mechanism, while the centralized methods can not do so.

\begin{figure}[ht]
    \centering
    \includegraphics[width=3.3in]{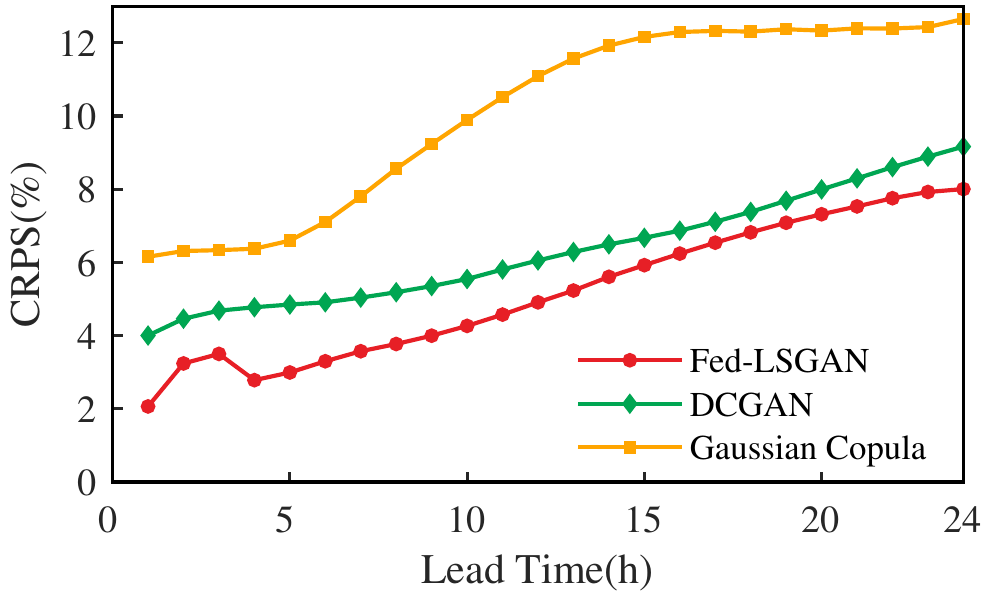}
    \caption{CRPS of scenarios generated by Fed-LSGAN and other methods.}
    \label{fig.crps}
\end{figure}

In order to further examine the quality of the scenarios generated by different methods, we adopt the well-known continuous ranked probability score (CRPS) which is a measure describing the dissimilarity of the cumulative distributions between generated scenarios and historical observations. The smaller the CRPS value, the closer the two cumulative distributions. The score at lead time $l$ is defined as
\begin{equation}
    \text{CRPS}_{l}=\frac{1}{M} \sum_{t=1}^{M}\int_{0}^{1}\left(\widehat{F}_{t+l \mid t}(\xi)-\mathbf{1}\left(\xi\geq \xi_{t+l}\right)\right)^{2} d\xi
\end{equation}

where $M$ is the total number of scenarios, $\widehat{F}_{t+l\mid t}(\xi)$ denotes the cumulative distribution function of normalized scenario, and $\mathbf{1}\left(\xi\geq \xi_{t+l}\right)$ is the indicator function for comparing scenarios and observation. As illustrated in Fig.~\ref{fig.crps}, the Fed-LSGAN is superior to other alternatives as our approach has smaller CRPS than the others with different lead times.

Besides the above analysis from the intuitive visualization, this study employs multiple quantitative evaluation metrics to measure the quality of the generated scenarios and lists the assessment scores of different methods in Table.~\ref{tab.score}. To further verify the superiority of our method, we add the original GAN for comparison, and conduct multiple simulation experiments to obtain the standard deviation, which can reflect the measures of dispersion of the experimental results.

Among them, the first three metrics are widely used for evaluating the samples generated by DGMs~\cite{xu2018empirical}, while the others including energy  score  (ES), mean absolute error (MAE), and root mean square error (RMSE) are classic scenario evaluation metrics~\cite{pinson2012evaluating}.

1) \emph{Fréchet Inception Distance (FID)}: FID is a powerful evaluation metric for DGMs, which is defined as
\begin{equation}
    \text{FID}({P_{d}},{P_{g}})=\left\| {\mu_{d}}-{\mu_{g}} \right\|+\text{Tr}({\Sigma_{d}}+{\Sigma_{g}}-2{{({\Sigma_{d}}{\Sigma_{g}})}^{\frac{1}{2}}})
\end{equation}
where $\mu_{d}$ and $\mu_{g}$ represent the empirical mean; $\Sigma_{d}$ and $\Sigma_g$ are empirical covariance. The smaller the value of FID, the closer the two distributions, which means the better the performance of the DGMs. 

2) \emph{Kernel Maximum Mean Discrepancy (MMD)}: MMD measures the difference between $P_{d}$ and $P_g$ for some fixed kernel funciton $k$, which is defined as
\begin{equation}
    \text{MMD}^2(P_{d},P_g)=\mathbb{E}_{\substack{x,x'\sim P_{d}\\y,y'\sim P_g}}\left[k(x,x')-2k(x,y)+k(y,y')\right]
\end{equation}
A lower KMMD suggests that the $P_{d}$ is closer to $P_g$.

3) \emph{The 1-Nearest Neighbor classifier (1-NN)}: It is used in two-sample tests to assess whether two distributions are identical. Given two sets of samples $S_d\sim P_{d}$ and $S_g\sim P_g$ with $|S_d|=|S_g|$, we can compute the leave-one-out (LOO) accuracy of a 1-NN classifier trained on $S_d$ and $S_g$ with positive labels for $S_d$ and negative labels for $S_g$. The LOO accuracy of the 1-NN classifier should eventually fluctuate around 50\%, which indicates that the two distributions match.

4) \emph{Energy Score (ES)}: This metric measures the deviation between the probability density of the renewable energy output scenarios and the observed results. The formula for the ES is
\begin{equation}
    \text{ES}=\frac{1}{M}\sum_{i=1}^M\,||\varsigma -\xi_i||-\frac{1}{2{{M}^{2}}}\sum_{i=1}^M\,\sum_{j=1}^M\,||\xi_i-\xi_j||
\end{equation}
where $\varsigma$ is the real renewable power output, $\xi_i$ is the $i$th generated time series scenario, and $M$ denotes the number of scenarios. A smaller $\text{ES}$ suggests smaller differences between the generated and actual scenarios.

Table.~\ref{tab.score} shows that the Fed-LSGAN can achieve excellent performances and outperforms other centralized DGM methods and Gaussian-copula method while ensuring data privacy and low communication overheads. More importantly, compared with the other alternatives, the Fed-LSGAN has basically smaller standard deviations in all metrics, which suggests that our approach has better stability in most cases.
\begin{table}
    \centering
    \scriptsize
    \caption{The scores(Standard Deviation) of Fed-LSGAN, DCGAN, GAN and Gaussian copula methods by different evaluation metrics.}
    \label{tab.score}
    \begin{tabular}{@{}ccccc@{}}
    \toprule
    Method               & Fed-LSGAN & DCGAN  & GAN & Copula \\ 
    \midrule
    FID                 & 4.446 (0.392)   & 4.879 (0.596)   & 7.057 (0.744)   &  -     \\
    MMD                 & 0.071 (0.006)   & 0.101 (0.014)   & 0.194 (0.013)   &  -     \\
    1-NN                & 0.525 (0.064)   & 0.536 (0.054)  & 0.895 (0.072)   &  -     \\
    ES                  & 0.940 (0.018)   & 1.084 (0.053)   & 2.289 (0.083)   & 1.146 (0.014)  \\
    MAE                 & 0.170 (0.015)   & 0.208 (0.014)   & 0.316 (0.045)   & 0.301 (0.020) \\
    RMSE                & 0.246 (0.029)   & 0.286 (0.015)  & 0.374 (0.051)    & 0.385 (0.065) \\
    \bottomrule
    \end{tabular}
\end{table}

\subsection{Spatial Correlation Analysis of Global Model}
As a large-scale system, a power system has a large number of renewable energy sites, so the renewable energy outputs between different locations show a significant spatial correlation due to the interdependence of wind speed or solar irradiance activities~\cite{papaefthymiou2008using}. Moreover, considering the spatial correlation in a scheduling model is beneficial to allocate the reserve capacity geographically to ensure the proper delivery of reserve power under congested transmission conditions~\cite{zhang2013modeling}. Unfortunately, traditional model-based methods are very troublesome when dealing with the spatial correlation. In this study, the global model trained by the combination of the proposed Fed-LSGAN and conditional generation~\cite{mirza2014conditional} can capture spatial correlations between all renewable sites that participate in the calculation. This work sets a client number as the label for all samples of each client.

To analyze the spatial relationship between multiple sites, we use the Pearson correlation coefficient to represent the correlation between different renewable sites. The Pearson correlation coefficient $\rho$ of two time series $S_i$ and $S_j$ is 
\begin{equation}
    \rho(S_i,S_j)=\frac{\sum_{i=1}^{n}(S_i-\bar{S_i})(S_j-\bar{S_j})}{\sqrt{\sum_{i=1}^{n}(S_i-\bar{S_i})^2}\sqrt{\sum_{i=1}^{n}(S_j-\bar{S_j})^2}} 
\end{equation}

\begin{figure}
    \centering
    \includegraphics[width=3.4in]{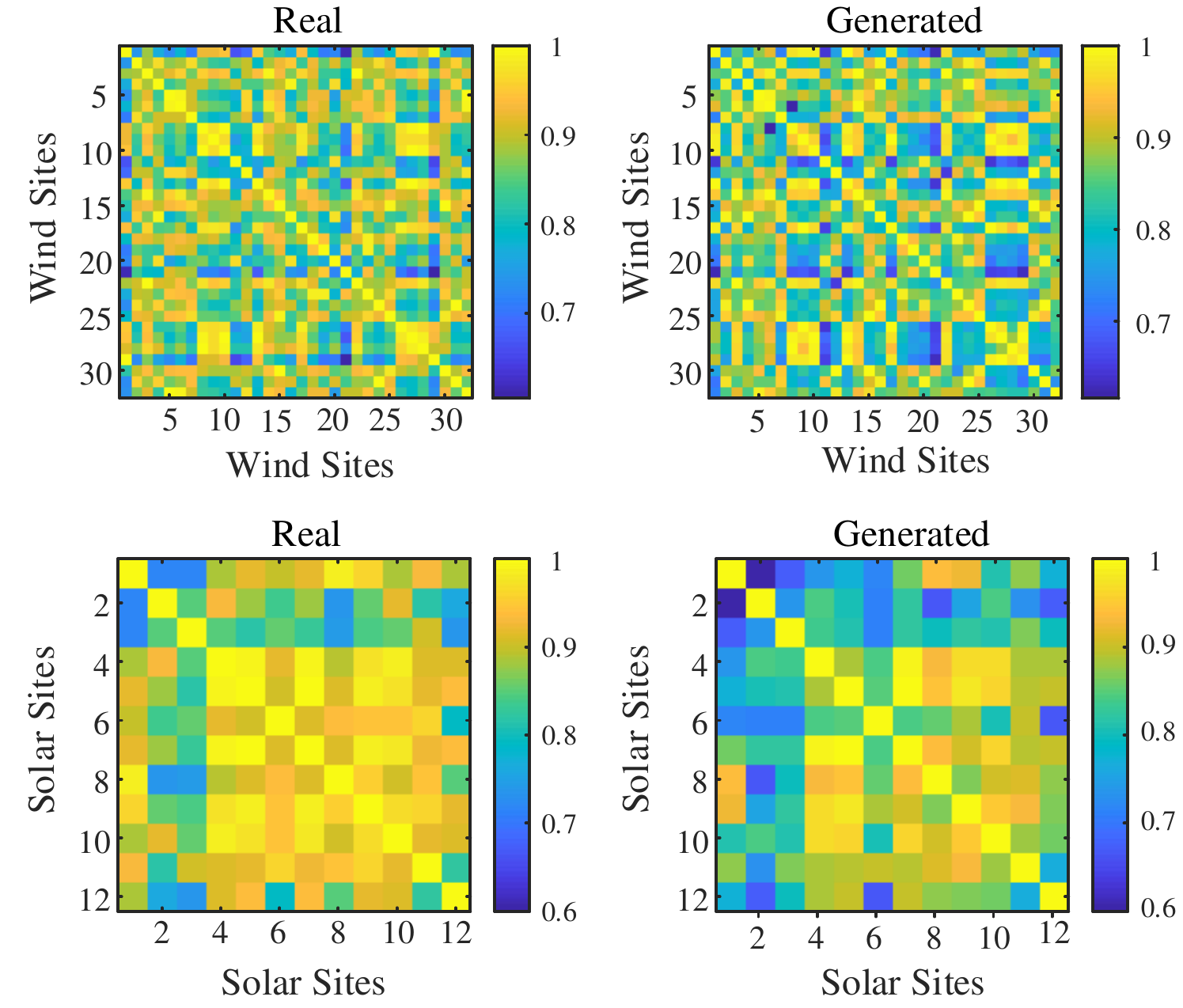}
    \caption{Multi-sites spatial correlation coefficient matrix colormap.}
    \label{fig.colormap}
\end{figure}

By using the global model to generate scenarios of all renewable sites, we calculate the Pearson correlation coefficient between the power scenarios of different sites, and then visualize them in Fig.~\ref{fig.colormap}. This figure depicts that the correlation between some sites is of the overall conformity to the correlation law of real data, although it is slightly reduced compared with the real samples. This illustrates that the Fed-LSGAN can capture not only the temporal characteristics of renewable energies but also the spatial characteristics among multiple sites. More importantly, it can preserve data privacy of each renewable power plant and reduce communication overheads when training the model.

\subsection{Robustness Analysis of Fed-LSGAN to Federated Settings}
\begin{figure}
    \centering
    \includegraphics[width=3.4in]{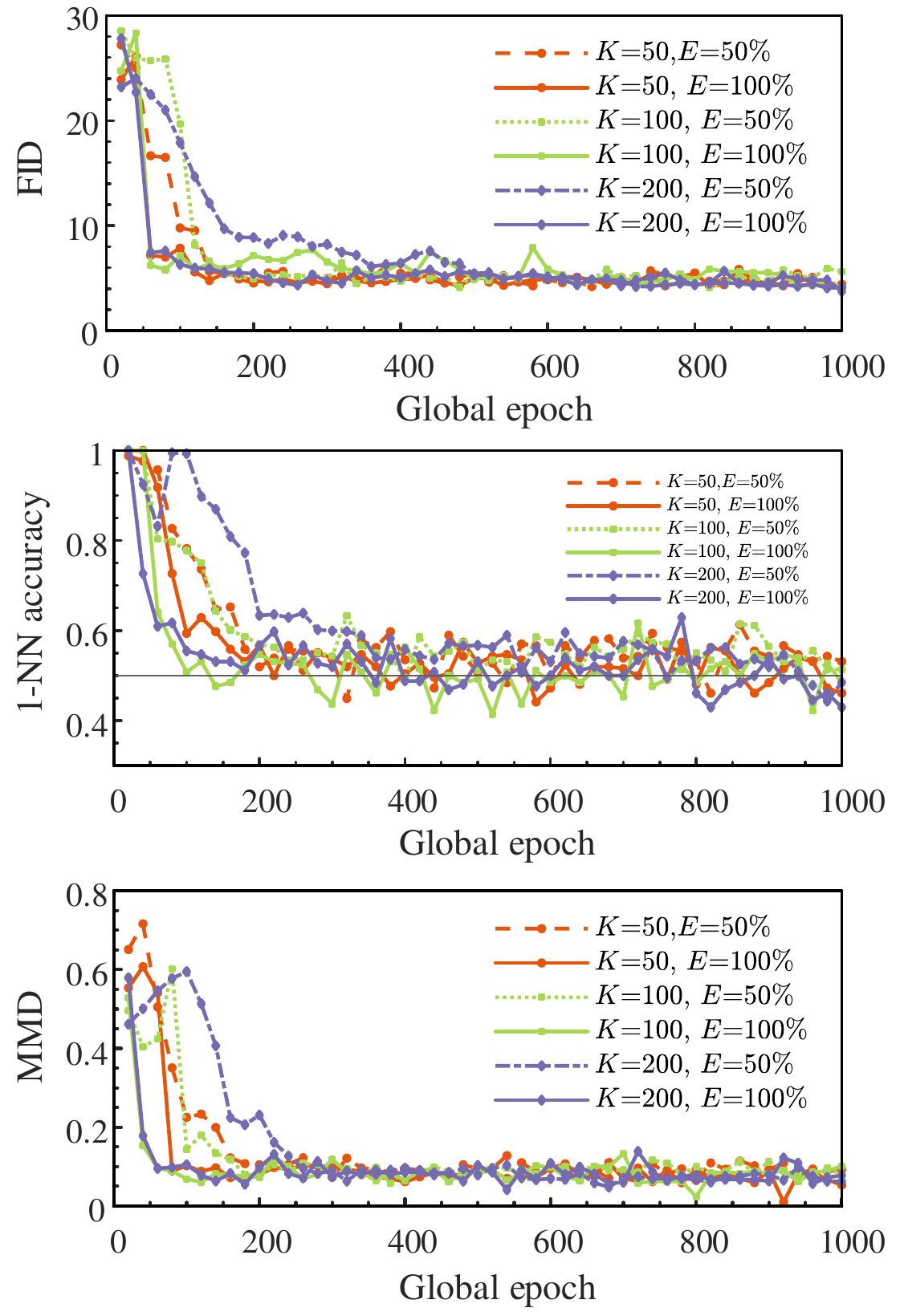}
    \caption{The simulations results with different federated setting.}
    \label{fig.score}
\end{figure}

To analyze the impact of the main control parameters on the robustness of the proposed Fed-LSGAN, we designed an experiment with varied synchronization intervals $K$ and the fractions $E$. Here, the synchronization intervals are respectively taken as 50, 100 and 200, while the fractions are set to 50\% and 100\%. The simulations results are shown in Fig.~\ref{fig.score}.

Fig.~\ref{fig.score} illustrates that different federated settings have different effects on the performances of the Fed-LSGAN. To be specific, for synchronization intervals, these metric curves suggest that the our approach's performance is basically unchanged with different intervals when the fraction is fixed, which proves that the Fed-LSGAN has good robustness to different synchronization intervals; regarding the fractions, the curves illustrate that different fractions do not have an effect on the scenario generation quality of the trained model, but affects the converge speed of the proposal for a given interval.  

\section{Conclusion}\label{sec.conclusion}
Scenario generation has been regarded as a fundamental and crucial tool for decision-making in power systems with high-penetration renewables due to its powerful ability to model the uncertainties of renewable power generations. This work presents Fed-LSGAN, an innovative federated deep generative learning framework, for renewable scenario generation. To the best of the authors' knowledge, this is the first attempt that leverages federated learning for scenario generation. Based on huge amounts of historical data, the Fed-LSGAN takes the advantages of federated learning and deep generative models for renewable scenario generation. Here, federated learning enables the proposal to generate scenarios in a privacy-preserving manner without sacrificing the generation quality by transferring model parameters, rather than all data; while the LSGANs-based deep generative model is able to generate scenarios that conform to the distribution of historical data through capturing the spatial-temporal characteristics of renewable powers.

The case studies demonstrate that the Fed-LSGAN manages to perform renewable scenario generation; and furthermore, the qualitative and quantitative evaluation results prove that the proposal outperforms the state-of-the-art centralized methods. Besides, an experiment with different federated learning settings is designed and performed in this study, and the experiment results verify the robustness of our method.

As a data-driven approach, the presented methodology is applicable to stochastic processes of interest in electrical energy systems with high penetration of renewable energies. This paper hasn't considered the data packet loss and communication delay in communication modeling, while more realistic scenarios need take into account such cases. Another interesting topic is to incorporate this study for the planning and optimal operation of an electrical system with renewables.

\bibliographystyle{IEEEtran}
\bibliography{ref}

\end{document}